\documentclass[conference]{IEEEtran}
\IEEEoverridecommandlockouts
\usepackage{cite}
\usepackage{amsmath,amssymb,amsfonts}
\usepackage[linesnumbered, boxed, ruled]{algorithm2e}

\usepackage{booktabs}
\usepackage{multirow}
\usepackage{graphicx}
\usepackage{textcomp}
\usepackage{xcolor}
\usepackage{pifont}
\usepackage{multirow}
\usepackage{xspace}
\usepackage{caption}
\usepackage{enumitem}
\usepackage{wrapfig,lipsum}
\usepackage[colorlinks, linkcolor=red,  anchorcolor=blue, citecolor=blue]{hyperref}
\def\BibTeX{{\rm B\kern-.05em{\sc i\kern-.025em b}\kern-.08em
    T\kern-.1667em\lower.7ex\hbox{E}\kern-.125emX}}
\begin{document}

\newcommand{\cmark}{\ding{51}\xspace}%
\newcommand{\cmarkg}{\textcolor{lightgray}{\ding{51}}\xspace}%
\newcommand{\xmark}{\ding{55}\xspace}%
\newcommand{\xmarkg}{\textcolor{lightgray}{\ding{55}}\xspace}%

\newcommand{\KW}[1]{{\color{blue}{#1}}}
\newcommand{\FY}[1]{{\color{orange}{#1}}}
\newcommand{\SQ}[1]{{\color{cyan}{#1}}}
\newcommand{\JW}[1]{{\color{magenta}{\bf JW: #1}}}

\definecolor{cyan}{cmyk}{.3,0,0,0}

\newcommand{\RED}[1]{\textcolor{red}{#1}}
\newcommand{\YELLOW}[1]{\textcolor{yellow}{#1}}
\newcommand{\BLUE}[1]{\textcolor{blue}{#1}}
\definecolor{springgreen}{RGB}{0,255,0}

\newcommand{\lsimple}{L_{DM}}

\newcommand{\model}{\epsilon_\theta}
\newcommand{\modeluncond}{\tilde{\epsilon}_\theta}

\newcommand{\conditioner}{\tau_\theta}
\newcommand{\expec}{\mathbb{E}}
\newcommand{\encoder}{\mathcal{E}}
\newcommand{\decoder}{\mathcal{D}}

\newcommand{\textprompt}{\mathcal{P}}
\newcommand{\textembedding}{\mathcal{C}}

\newcommand{\updateprompt}{\hat{\mathcal{C}_t}}
\newcommand{\tokenset}{\mathcal{V}}
\newcommand{\updateset}{{\mathcal{V}_t}}
\newcommand{\updatetoken}{{v_t^k}}
\newcommand{\updateTKsubi}{{v_t^i}}
\newcommand{\updateTKsubj}{{v_t^j}}
\newcommand{\updateTK}{{v_t}}

\newcommand{\background}{\mathcal{B}}
\newcommand{\crossattn}{\mathcal{A}}
\newcommand{\gaussian}{\mathcal{F}}
\newcommand{\cluster}{\mathcal{M}}
\newcommand{\mlp}{\mathit{l}}

\newcommand{\ddimz}{z}
\newcommand{\denoisez}{{\bar z}}
\newcommand{\interz}{{\tilde z}}

\newcommand{\inputimage}{\mathcal{I}}

\newcommand{\tabincell}[2]{\begin{tabular}{@{}#1@{}}#2\end{tabular}}
\newcommand{\minisection}[1]{\vspace{0.01in} \noindent {\bf #1}\ }

\title{IterInv: Iterative Inversion for Pixel-Level T2I Models} 


\author{\IEEEauthorblockN{1\textsuperscript{st} Chuanming Tang}
\IEEEauthorblockA{\textit{University of Chinese Academy of Sciences} \\
\textit{Institute of Optics and Electronics} \\
\textit{Computer Vision Center}\\
Beijing, China \\
tangchuanming19@mails.ucas.ac.cn}
\and
\IEEEauthorblockN{2\textsuperscript{nd} Kai Wang*\thanks{*Corresponding author}}
\IEEEauthorblockA{\textit{Computer Vision Center} \\
Barcelona, Spain \\
kwang@cvc.uab.es}
\and
\IEEEauthorblockN{3\textsuperscript{rd} Joost van de Weijer }
\IEEEauthorblockA{\textit{Computer Vision Center} \\
\textit{Universitat Autonoma de Barcelona}\\
Barcelona, Spain \\
joost@cvc.uab.es}
}

\maketitle

\begin{abstract}
Large-scale text-to-image diffusion models have been a ground-breaking development in generating convincing images following an input text prompt. The goal of image editing research is to give users control over the generated images by modifying the text prompt. Current image editing techniques predominantly hinge on DDIM inversion as a  prevalent practice rooted in Latent Diffusion Models (LDM). However, the large pretrained T2I models working on the latent space suffer from losing details due to the first compression stage with an autoencoder mechanism. Instead, other mainstream T2I pipeline working on the pixel level, such as Imagen and DeepFloyd-IF, circumvents the above problem. They are commonly composed of multiple stages, typically starting with a text-to-image stage and followed by several super-resolution stages. In this pipeline, the DDIM inversion fails to find the initial noise and generate the original image given that the super-resolution diffusion models are not compatible with the DDIM technique. According to our experimental findings, iteratively concatenating the noisy image as the condition is the root of this problem. Based on this observation, we develop an iterative inversion (IterInv) technique for this category of T2I models and verify IterInv with the open-source DeepFloyd-IF model.
Specifically, IterInv employ NTI as the inversion and reconstruction of low-resolution image generation. In stages 2 and 3, we update the latent variance at each timestep to find the deterministic inversion trace and promote the reconstruction process. 
By combining our method with a popular image editing method, we prove the application prospects of IterInv. The code will be released upon acceptance. 
The code is available at \href{https://github.com/Tchuanm/IterInv.git}{https://github.com/Tchuanm/IterInv.git}
\end{abstract}

\begin{IEEEkeywords}
 Image Inversion, Image Reconstruction, Image Editing, Text-to-Image, Pixel Diffusion
\end{IEEEkeywords}

\section{Introduction}
The Text-to-Image (T2I) field has witnessed significant advancements and demonstrated an unprecedented ability to generate realistic images~\cite{midjourney,ramesh2021zero,Rombach_2022_CVPR_stablediffusion,saharia2022imagen,deepfloyd}. 
The pioneering text-to-image frameworks based on diffusion model can be roughly categorized considering where the diffusion prior is conducted, i.e., the pixel space or latent space. The first class of methods generate images directly from the high-dimensional pixel level, including GLIDE~\cite{GLIDE}, Imagen~\cite{saharia2022imagen} and DeepFloyd-IF~\cite{deepfloyd}. Another stream of works proposes to first compress the image to a low-dimensional space, and then train the diffusion model on this latent space. Representative methods falling into this category include LDM~\cite{Rombach_2022_CVPR_stablediffusion} and DALL-E~\cite{ramesh2022dalle2}. 
State-of-the-art T2I models undergo training on immensely large language-image datasets, demanding substantial computational resources. 
Nevertheless, notwithstanding their remarkable abilities, these models do not readily facilitate \textit{real image editing}.

\textit{Text-guided image editing}, also referred to as prompt-based image editing, empowers users to effortlessly modify an image exclusively through text prompts~\cite{cao2023masactrl,couairon2023diffedit,hertz2023delta_DDS,hertz2022prompt,li2023stylediffusion,tumanyan2022plug,wang2023mdp}. 
arious methods~\cite{bar2022text2live,choi2021ilvr,Kim_2022_CVPR,kwon2023diffusionbased,li2023stylediffusion} leverage CLIP~\cite{radford2021clip} for image editing based on a pretrained \textit{unconditional} diffusion model.
However, they are limited to the generative prior which is only learned from visual data of a specific domain, whereas the CLIP text-image alignment information is from much broader data. This prior gap is mitigated by recent progress of T2I models~\cite{chang2023muse,gafni2022make,hong2022sag,ramesh2022dalle2,ramesh2021zero,saharia2022imagen}. Nevertheless, these T2I models offer little control over the generated contents. This creates a great interest in developing methods~\cite{brooks2022instructpix2pix,hertz2022prompt,kawar2022imagic,meng2022sdedit,parmar2023zero,patashnik2023localizing,tumanyan2022plug,kai2023DPL} to adopt such T2I models for controllable image editing.  However, how to incorporate these methods with pixel-level T2I models is still an open topic, and we are the first to shed light on this direction.

\begin{figure*}[thb]
\centering
\includegraphics[width=0.99\linewidth]{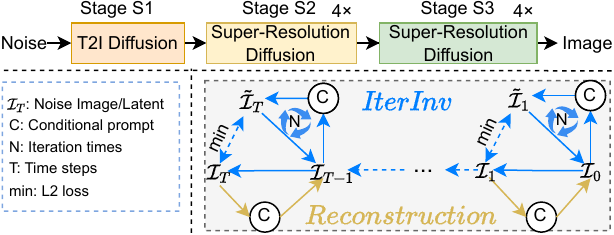}
\caption{The network of DeepFloyd-IF pipeline and proposed IterInv inversion technology.}
\label{fig:network}
\end{figure*}

Several of the existing methods use DDIM inversion~\cite{song2021ddim} as a common practice to attain the initial latent code of the image and then apply their proposed editing techniques along the denoising phase~\cite{mokady2022null,parmar2023zero,tumanyan2022plug}. 
DDIM inversion~\cite{song2021ddim} shows significant potential in editing tasks by deterministically calculating and encoding context information into a latent space and then reconstructing the original image using this latent representation. 
However, DDIM is found lacking for text-guided diffusion models when classifier-free guidance (CFG)~\cite{ho2022classifier} is applied, which is necessary for meaningful editing.
Leveraging optimization on Null-Text embedding, Null-Text Inversion (NTI)~\cite{mokady2022null} further improved the image reconstruction quality when CFG is applied, and retained the rich text-guided editing capabilities of the Stable Diffusion model~\cite{Rombach_2022_CVPR_stablediffusion}. 
 Negative-prompt inversion (NPI)~\cite{miyake2023NPI} and ProxNPI~\cite{han2023ProxNPI} further reduces the computation cost for the inversion step while generates similarly competitive reconstruction results as Null-Text inversion. 
Nevertheless, current text-guided editing methods are mainly working on latent-level T2I Latent Diffusion Models (LDM)~\cite{Rombach_2022_CVPR_stablediffusion}. And the first compression stage with an Autoencoder results in losing details of the original images as our experiments show. 
Instead, another mainstream of pixel-level T2I diffusion models~\cite{saharia2022imagen,deepfloyd} avoids this problem by directly generating images over the pixel space. However, the inversion and text-guided image editing techniques are not well explored for this branch.

In this paper, we take advantage of the open-source DeepFloyd-IF model as the representative for pixel-level T2I models.
We first observe that directly applying the DDIM inversion to the open-source T2I DeepFloyd-IF model will lead to failures in reconstructing the original images. We attribute this phenomenon to the concatenation conditioning with the noisy images in the super-resolution diffusion models.
To solve this problem, we propose the iterative inversion  (IterInv) technique, where we find the trace back in the diffusion process by iterative optimization to approximate the real image.

From our experiments over public datasets collected from previous text-guided editing papers~\cite{parmar2023zero,tumanyan2022plug}, we verify the extraordinary reconstruction ability of IterInv and successfully combine it with DiffEdit~\cite{couairon2023diffedit} to show its compatibility with editing methods. We prospect this proposal can help to innovate future research on text-guided image editing based on the pixel-level T2I models.

\section{Methodology}
In this section, we first provide a short preliminary and then describe our method IterInv in detail to cooperate with the image editing method with pixel-level T2I medel demand. 
To our knowledge, our IterInv is the first method to solve the DDIM inversion drawback in the cascaded diffusion pipeline and reconstruct the original at the pixel-level.

\subsection{Preliminary}

\minisection{DeepFloyd-IF.} We develop our method for the publicly available DeepFloyd-IF method, but the method is general and can be extended to other diffusion models. Pixel-level T2I DeepFloyd-IF~\cite{deepfloyd} is a cascaded diffusion model. 
Different from the current popular latent model SD~\cite{Rombach_2022_CVPR_stablediffusion} which consists of a single-stage diffusion phase, 
it is composed of three cascaded pixel-level diffusion stages: a text-based low-resolution text-to-image generation stage S1, and two super-resolution diffusion modules as stage S2 and S3. The output resolutions of these stages are 64, 256, and 1024 respectively.
Compared with SD, DeepFloyd-IF contains two advantages. In stage I, DeepFloyd-IF generates an image without projecting the image into latent space and with less calculation cost.  Further, the super-resolution module of DeepFloyd-IF can synthesize higher resolution with pixel details compared with SD ($1024~vs.~512$). 
In stage S1, given the conditional input prompt $\textprompt$ and a random noise  $\mathcal{I}^{s_1}_T$, 
a conditioning mechanism denoted as $\tau_\theta(\textprompt)$ is employed, which maps the condition $\textprompt$ into a prompt vector.

\minisection{DDIM Inversion.} Inversion entails finding an initial  noise $z_T$ that reconstructs the code $z_0$ of the real image upon sampling. 
Existing image editing methods~\cite{brooks2022instructpix2pix,han2023ProxNPI} aim at precisely reconstructing a given image for editing, therefore employ the deterministic latent DDIM sampling~\cite{song2021ddim}:
\begin{equation}
 z_{t+1} = \sqrt{\bar{\alpha}_{t+1}}f_\theta(z_t,t,\textembedding) + \sqrt{1-\bar{\alpha}_{t+1}} \model(z_t,t,\textembedding). 
\end{equation}
$\bar{\alpha}_{t+1}$ is noise scaling factor defined in DDIM~\cite{song2021ddim} and $f_\theta(z_t,t,\textembedding)$ predicts the denoised latent code $z_0$ as:
\begin{equation}
f_\theta(z_t,t,\textembedding) = \Big[z_t - \sqrt{1-\bar{\alpha}_t} \model(z_t,t,\textembedding) \Big] / {\sqrt{\bar{\alpha}_t}}.
\end{equation}
However, with the pixel-level inversion, the concatenation condition with a noisy image in stages S2 and S3 of the DeepFloyd-IF model complicates the inversion process, and the generated image based on the found initial noise image of DDIM inversion deviates from the original image.

\minisection{Null-Text Inversion.}  
To amplify the effect of conditional textual input, classifier-free guidance~\cite{ho2022classifier} is proposed to extrapolate the conditioned noise prediction with an unconditional noise prediction. Let $\varnothing = \tau_\theta(\mathcal{`` "})$ denote the null text embedding, then the classifier-free guidance is defined as: 
\begin{equation}
\modeluncond(z_t,t,\textembedding,\varnothing) = w \cdot \model(z_t,t,\textembedding) + (1-w) \ \cdot \model(z_t,t,\varnothing)
\label{eq:classifier_free}
\end{equation}
where we set the guidance scale $w=7.5$ as is standard for Stable Diffusion~\cite{ho2022classifier,mokady2022null,Rombach_2022_CVPR_stablediffusion}.
However, the introduction of amplifier-free guidance complicates the inversion process, and the generated image based on the found initial noise $z_T$ deviates from the original image.
Null-Text inversion~\cite{mokady2022null} proposes a novel optimization which updates the null text embedding $\varnothing_t$ for each DDIM step $t \in [1,T]$ to approximate the DDIM trajectory $\{\ddimz_t\}_0^T$ according to: 
\begin{equation}
    \min_{\varnothing_t} \left \| {\denoisez}_{t-1}- \modeluncond( {\denoisez}_t,t,\textembedding,\varnothing_t) \right \|^2_2
\end{equation}
where $\{\denoisez_t\}_0^T$ is the backward trace from Null-Text inversion.

\subsection{IterInv: Iterative Inversion} 

DDIM inversion provides a deterministic process for conventional diffusion models~\cite{ho2020ddpm,song2021ddim}, similar to the first stage S1 of the DeepFloyd-IF model.
However, directly applying DDIM to DeepFloyd-IF leads to failure reconstructions as shown in Figure~\ref{fig:vis_appendix}. 
In the second and third columns, we only apply DDIM inversion with real images in the super-resolution stages S2 and S3. 
It is evident that the reconstructions retain only the layout structure, while other details are lost.
Then with the DDIM inversion image as an output of stage S1, the reconstruction is even worse as shown in the fourth column.
Based on this finding, we devise a novel iterative inversion framework (IterInv) that successfully injects the DDIM into the super-resolution stages S2 and S3.  The detail of the cascaded network is shown in Figure~\ref{fig:network} and the algorithm diagram is shown in Algorithm~\ref{alg:algorithm}.  
For each input image, we resize into three scales 64,256,1024 as inputs for different stages $x^{s_1},x^{s_2},x^{s_3}$.

\minisection{Stage S1.} In stage S1, with the low-resolution input image $x^{s_1}$, we employ Null-Text inversion~\cite{mokady2022null} in conjunction with classifier-free guidance~\cite{ho2022classifier} to approximate the original image.

\minisection{Stage S2/S3.} 
In stages S2 and S3,  the super-resolution diffusion models can operate at either pixel level (i.e., DeepFloyd-IF-upscaler) or latent level (i.e., SD-upscaler).
In stage S2, we use $\mathcal{I}^{s_2}_0$ to formulate the real image $x^{s_2}$ or latent code $z^{s_2}_0$. With the input as $\mathcal{I}^{s_2}_0$,  we design the reverse iterative inversion process for each timestep. 
Specifically, in the process of $\mathcal{I}^{s_2}_{t-1} \rightarrow \mathcal{I}^{s_2}_{t}$, we design a $N$ steps iterative optimization operation. 
We initialize $\tilde{\mathcal{I}}^{s_2}_t=\mathcal{I}^{s_2}_{t-1}$, and perform the following optimization for $N$ times iterations:
\begin{equation}  
\min ||\mathcal{I}^{s_2}_{t-1} - \modeluncond(\tilde{\mathcal{I}}^{s_2}_t, \hat{\mathcal{I}}^{s_1}_0,t,{\textembedding}) ||^2_2  
\end{equation}

Here $\modeluncond(\tilde{\mathcal{I}}^{s_2}_t,\hat{\mathcal{I}}^{s_1}_0,t,{\textembedding})$ is the DDIM sampling on $\tilde{\mathcal{I}}^{s_2}_t$ and the noised  reconstruction image $\hat{\mathcal{I}}^{s_1}_0$ from the previous stage.  After the optimization, we update the input as:
\begin{equation}
{\mathcal{I}}^{s_2}_{t} = \modeluncond(\tilde{\mathcal{I}}^{s_2}_t,\tilde{\mathcal{I}}^{s_1}_0, t, \textembedding).
\end{equation} 
Afterwards, with the $N$ iteration optimization of each timestep, we minimize the disparity between the predicted and excepted noise. Then, we regard the last iteration predict noise ${\mathcal{I}}^{s_2}_{t}$ as the trace of the inversion process of step $t$. 
The final noise ${\mathcal{I}}^{s_2}_T$ is the ideal trace for promoting the follow-up accurate image reconstruction and editing task. 
Stage S3 shares a similar mechanism as stage S2, and thus we employ the same IterInv strategy to obtain further high-resolution image reconstruction.

\begin{algorithm}[t]
\SetAlgoLined
\textbf{Input:} A source prompt $\textprompt$, \\
~~~~~~~~~~~~Three scale input images $x^{s_1}, x^{s_2}, x^{s_3}$ \\
\textbf{Initialize:} $\varnothing_T = \tau_\theta(\mathcal{``"})$, 
  ${\textembedding}=\tau_\theta({\textprompt});$ \\ ~~~~~~~~~~~~~~~$\omega_1=7.0,\omega_2=1.0,\omega_3=1.0;$ \\
  ~~~~~~~~~~~~~~~$ \mathcal{I}^{s_1}_0=x^{s_1}, \mathcal{I}^{s_2}_0 = x^{s_2}, \mathcal{I}^{s_3}_0 = x^{s_3}$; \\
  \textbf{Stage S1:} 
      $\mathcal{I}^{s_1}_{T},  \{\varnothing_t\}^T_{t=1}, \bar{\mathcal{I}}_0^{s_1} \gets \textbf{NTI}(\mathcal{I}^{s_1}_0, \textembedding, \varnothing_T)$ \\ 
\textbf{Stage S2:} \\
      $ \hat{\mathcal{I}}_0^{s_1} \gets \bar{\mathcal{I}}_0^{s_1} + noise $  \\
 \For{$t=1,2,\ldots,T$}{
    $\tilde{\mathcal{I}}^{s_2}_t \gets \mathcal{I}^{s_2}_{t-1};$ \\
        \For{$j=1,...,N$ }  
        {         
        $\tilde{\mathcal{I}}^{s_2}_{t-1} = \modeluncond(\tilde{\mathcal{I}}^{s_2}_t, \hat{\mathcal{I}}_0^{s_1}, t,{\textembedding})$; \\
        $L_t = ||\mathcal{I}^{s_2}_{t-1} - \tilde{\mathcal{I}}^{s_2}_{t-1} ||_2$; \\
        $ \tilde{\mathcal{I}}^{s_2}_t \gets  \tilde{\mathcal{I}}^{s_2}_t - \frac{\Delta L_t}{\Delta\tilde{\mathcal{I}}^{s_2}_t}$ \\
        }
    ${\mathcal{I}}^{s_2}_t \gets \tilde{\mathcal{I}}^{s_2}_t;$  
 }
$ \bar{\mathcal{I}}_0^{s_2} \gets ({\mathcal{I}}^{s_2}_T,\hat{\mathcal{I}}_0^{s_1},\textembedding)$ ~~~~~~~~~~~~~  \hfill  $\triangleleft$ \textit{Reconstruction}  \\
\textbf{Stage S3:} Similar as lines 8-18 computing  ${\mathcal{I}}^{s_3}_{T}$ \\
 \textbf{Return:} $
 \mathcal{I}^{s_1}_T, \mathcal{I}^{s_2}_T, \mathcal{I}^{s_3}_T
 $ \\ 
 \caption{ Iterative Inversion (IterInv) }
\label{alg:algorithm}
\end{algorithm}

\subsection{Combing with Image Editing Method}
\begin{figure*}[t]
  \centering
\includegraphics[width=0.95\textwidth]{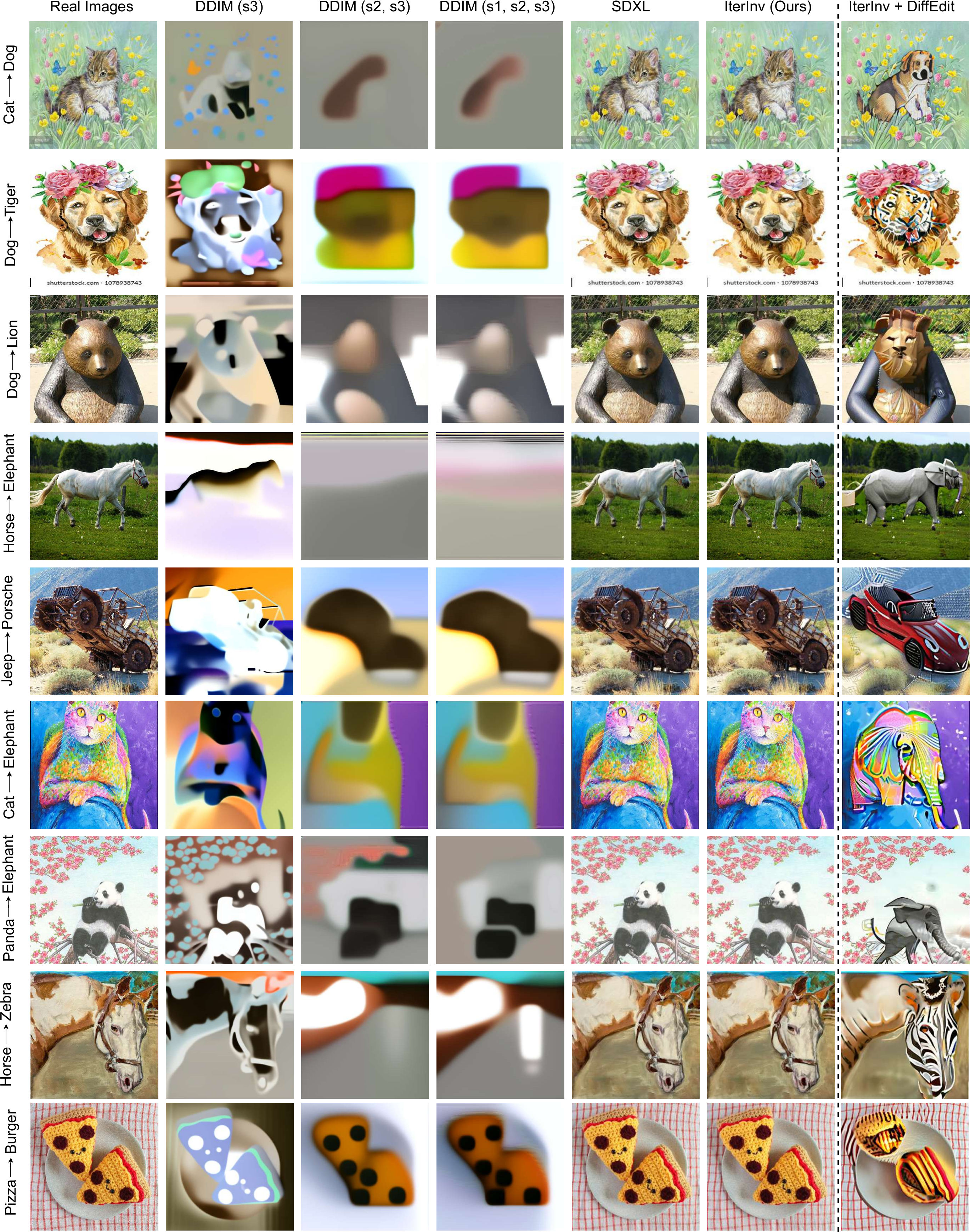}
  \caption{Visualization comparison of various inversion means and our editing results. } 
  \label{fig:vis_appendix}
\end{figure*}

DiffEdit~\cite{couairon2023diffedit} is an editing method of target implemented in SD manner, which highlights the target mask to edit based on different text prompts and relies on latent inference to preserve interest content region. 
It consists of three steps: 1) compute the target mask by estimating the noise conditionally. 2) encode with DDIM until encoding ratio \textit{r}. 3) decode and reconstruct the noise with mask-wise correction and inpainting strength. 
After successfully inverting real images with DeepFloyd-IF, we further enable the text editing technique on DeepFloyd-IF. To the best knowledge, our IterInv is the first to realize image editing with the pixel-level diffusion manner. With our deterministic iterative inversion, we introduce DiffEdit into DeepFloyd-IF to achieve text-guided image editing, as shown in the last column in Figure~\ref{fig:vis_appendix}.

\begin{table*}[t]
	\centering
	\newcommand{\best}[1]{\textbf{\textcolor{red}{#1}}}
	\newcommand{\scnd}[1]{\textbf{\textcolor{blue}{#1}}}
	\newcommand{\dist}{\hspace{4pt}}%
	\resizebox{1\linewidth}{!}{%
        \begin{tabular}{l@{\dist}|l@{\dist}|c@{\dist}|c@{\dist}|c@{\dist}|c@{\dist}|c@{\dist}|c@{\dist}}
        	\toprule
        Model & Method   &  $\omega_1$   & MSE ($\downarrow$) & LPIPS ($\downarrow$)    &   SSIM ($\uparrow$)      &PSNR ($\uparrow$)      & CLIP ($\uparrow$)    \\
        	\midrule
        SDXL & Autoencoder  &-  &0.009016    &0.1269     & 0.9048    &33.4993    &\textbf{21.3172}  \\
        \midrule
         \multirow{7}{*}{\rotatebox[origin=c]{90}{DeepFloyd-IF}} &    DDIM (s3)   &-   & 0.275662      &  0.7882     & 0.4213       & 6.8092      &   21.2448        \\
         &    DDIM (s2,s3)  &-    &0.076924  &0.6413  &0.5865   &11.8044 &21.1808 \\
         &    DDIM (s1,s2,s3)  & 1.0    & 0.079924      & 0.6393      &  0.5821     &  11.8366     &    21.2223
           \\
        	\cline{2-8}
           
        &    \multirow{4}{*}{{Ours (IterInv)}}   & 1.0  & 0.000130      & 0.0356
      & 0.9806      &  40.6102     &    21.2907          \\
              &&3.0   & 0.000129      &  0.0353    &  0.9806     & 40.6459      &   21.3161          \\
              &&5.0   & 0.000129      & 0.0353      &  0.9806     &  40.6484     &  21.3161          \\
              &&7.0   & \textbf{0.000129}       & \textbf{0.0353}      & \textbf{0.9806}      & \textbf{40.6484}      &   {21.3161}          \\
            \bottomrule
        \end{tabular}
	}
	\caption{ Evaluation of reconstruction quality of each method on \emph{ImageInversion} dataset.  } 
	\label{tab:inversion_evaluate}
\end{table*}

\section{Experiments}

\subsection{Data and Implementation}
We evaluate the proposed IterInv qualitatively and quantitatively. We experimented IterInv using DeepFloyd-IF with PyTorch and diffusers~\cite{von-platen-etal-2022-diffusers}. 
For quantitative comparison, we collect the \emph{ImageInversion} dataset with 69 images: 50 images from PnP~\cite{tumanyan2022plug} and 19 images from Pix2pix-zero~\cite{parmar2023zero}.
To ensure a fair comparison among different methods, all images are resized to $1024\times1024$.
To comprehensively evaluate the quality of inversion methods, we employ five metrics including MSE, LPIPS~\cite{zhang2018lpips}, SSIM~\cite{wang2003ssim}, PSNR, and CLIP score~\cite{hessel2021clipscore} to verify the effectiveness of IterInv. In IterInv, we set $T=50, N=20$ by default.

\subsection{Reconstruction and Editing}
The qualitative and quantitative measurements of reconstructions are shown in Figure~\ref{fig:vis_appendix} and Table~\ref{tab:inversion_evaluate} respectively. 
Compared with DDIM inversion applied in various stages and the \textit{first stage autoencoder }of the SDXL model, IterInv achieves more precise image construction and there is only the SDXL \textit{compression autoencoder}\footnote{The first stage compression model makes the upper bound for any inversion techniques based on the SDXL.} can reconstruct the target image closer to our final performance, as indicated by the five metric evaluations. 
Furthermore, we visualize the image editing results by applying  DiffEdit to IterInv in the last column of Figure~\ref{fig:vis_appendix}. With these convincing image editing results, we are showing the application prospect of IterInv combined with text-guided image editing methods in future research.

In Table~\ref{tab:inversion_evaluate}, MSE, LPIPS, SSIM, and PSNR all measure the quality of the original image with a reconstruction image, reflecting the inversion quality of different methods. However, the CLIP score is a method to measure an image's proximity to the provided prompt, which is not really related to the reconstruction quality compared with the original image. 
In Figure~\ref{fig:vis_appendix}, the reconstruction quality of SDXL Autoencoder and ours IterInv all looks good. However, as compared in Table~\ref{tab:inversion_evaluate}, the MSE, LPIPS, SSIM, and PSNR of our IterInv all outperform SDXL with a significant margin, which shows the excellent reconstruction ability of our IterInv in the pixel-level diffusion manner.

\subsection{Ablation study}
Current inversion methods on the T2I model are easily influenced by the hyperparameter of classifier-free guidance (CFG~\cite{mokady2022null}). We conduct the ablation study over various CFG values of stage S1 ($\omega_1$), as shown in Table~\ref{tab:inversion_evaluate}. 
With different values of $\omega_1$, IterInv achieves nearly the same reconstruction performance, which shows the parameter robustness of IterInv to the classifier-guidance scale.

\section{Conclusions}

Current text-guided image editing methods predominantly leverages latent diffusion models (LDMs), which often result in detail loss due to the low resolution of latent space representations. To circumvent this limitation, pixel-level T2I diffusion models, such as DeepFloyd-IF, offer a promising alternative. 
However, challenges persist with the widely used DDIM inversion technique, particularly its inability to ensure accurate reconstruction of real images when applied to the pixel-level DeepFloyd-IF model. This issue is largely attributed to the manner in which conditions are concatenated with noisy inputs within the model framework.
To address this limitation, we introduce Iterative Inversion (IterInv), a novel inversion technique designed specifically for this context. 
Through experimentation with real images, IterInv has demonstrated enhanced performance in both inversion accuracy and subsequent editing capabilities compared to traditional methods. 

\minisection{Limitations.}
Despite the thorough evaluation of our proposed method on real image inversion and editing tasks, it exhibits certain limitations. Primarily, our investigation in this paper is confined to the inversion problem utilizing the open-source DeepFloyd model. This restriction narrows the application scope of our approach, dubbed IterInv. Additionally, the inversion method has only been tested in conjunction with the DiffEdit approach for real image editing. Its compatibility and effectiveness with alternative editing methodologies remain unexplored, suggesting potential avenues for further research to broaden its applicability.

\section*{Acknowledgment}
We acknowledge projects TED2021-132513B-I00 and PID2022-143257NB-I00, financed by MCIN/AEI/10.13039/501100011033 and FSE+ and the Generalitat de Catalunya CERCA Program. 
Chuanming thanks the Chinese Scholarship Council (CSC) No.202204910331.



\clearpage

{\small
\bibliographystyle{IEEEtran}
\bibliography{longstrings,mybib}
}

\end{document}